\pdfoutput=1

\documentclass[11pt]{article}

\usepackage[preprint]{acl}

\usepackage{times}
\usepackage{latexsym}
\usepackage{booktabs}
\usepackage{subcaption}

\usepackage{tikz}
\usepackage{pgfplots}
\pgfplotsset{compat=1.18}

\usepackage[T1]{fontenc}

\usepackage[utf8]{inputenc}

\usepackage{microtype}

\usepackage{inconsolata}

\usepackage{graphicx}
\usepackage[most]{tcolorbox}
\usepgfplotslibrary{groupplots}
\usetikzlibrary{patterns}
\title{
Retrieval Augmented Question Answering: \\ When Should LLMs Admit Ignorance?
}

\author{
  Dingmin Wang\textsuperscript{\rm 1,2}\Thanks{Work done as a student researcher at Google Research.} \quad Ji Ma\textsuperscript{\rm 1} \quad Shankar Kumar\textsuperscript{\rm 1}
  \\
  \textsuperscript{\rm 1}Google Research \qquad 
  \textsuperscript{\rm 2}University of Oxford 
  \\
  \texttt{dingmin.wang@cs.ox.ac.uk}, \texttt{\{maji,shankarkumar\}@google.com}
}

\begin{document}
\maketitle
\begin{abstract}
The success of expanded context windows in Large Language Models (LLMs) has driven increased use of broader context in retrieval-augmented generation. We investigate the use of LLMs for retrieval augmented question answering. While longer contexts make it easier to incorporate targeted knowledge, they introduce more irrelevant information that hinders the model's generation process and degrades its performance. To address the issue, we design an adaptive prompting strategy which involves splitting the retrieved information into smaller chunks and sequentially prompting a LLM to answer the question using each chunk.  Adjusting the chunk size allows a trade-off between incorporating relevant information and reducing irrelevant information.  Experimental results on three open-domain question answering datasets demonstrate that the adaptive strategy matches the performance of standard prompting while using fewer tokens. Our analysis reveals that when encountering insufficient information, the LLM often generates incorrect answers instead of declining to respond, which constitutes a major source of error. This finding highlights the need for further research into enhancing LLMs' ability to effectively decline requests when faced with inadequate information.

\end{abstract}

\section{Introduction}

Recent research on large language models (LLMs) has shown significant progress in handling very long context lengths \cite{achiam2023gpt, anthropic2024claude, beltagy2020longformer, guo2021longt5, glm2024chatglm}, with Gemini 1.5 \cite{reid2024gemini} capable of processing 2 million tokens, enabling new capabilities such as in-context learning from an entire book.

Retrieval Augmented Generation (RAG)~\cite{guu2020retrieval, lewis2020retrieval, gao2024ragsurvay} augments Large Language Models (LLMs) by extracting pertinent information from external databases (e.g., Wikipedia) and incorporating it into the model's context window. This process guides the model towards generating grounded and accurate responses. While expanding the context window theoretically improves the recall of relevant information, it does not guarantee higher accuracy. Longer contexts increase the probability of including irrelevant information (distractors). Recent research indicates that the accumulation of irrelevant data can disrupt the generation process and degrade the quality of the model's output~\cite{hsieh2024ruler, lee2024longcontextlanguagemodelssubsume,wang2024retrieve}.

In this work, we investigate the efficacy of a divide-and-conquer approach for prompting within RAG frameworks. We propose an adaptive prompting strategy that partitions long contexts into smaller, manageable windows, thereby decomposing the retrieval task into a series of sub-problems. Specifically, if a window yields sufficient information, the LLM generates the answer and terminates the process; otherwise, it signals the insufficiency and iterates to the next window. This methodology presents two distinct advantages:
\begin{description}
\item [Mitigation of Noise:]By isolating segments of text, we reduce the accumulation of irrelevant information within the active context. This lowers the reasoning complexity for the model, allowing it to focus on a cleaner signal rather than requiring it to filter a surplus of noise.
\item [Extended Capabilities:]The sequential processing circumvents standard context length limitations, resulting in improved recall.
\end{description}

\begin{figure*}[!htp]
    \centering
    \includegraphics[width=0.95\linewidth]{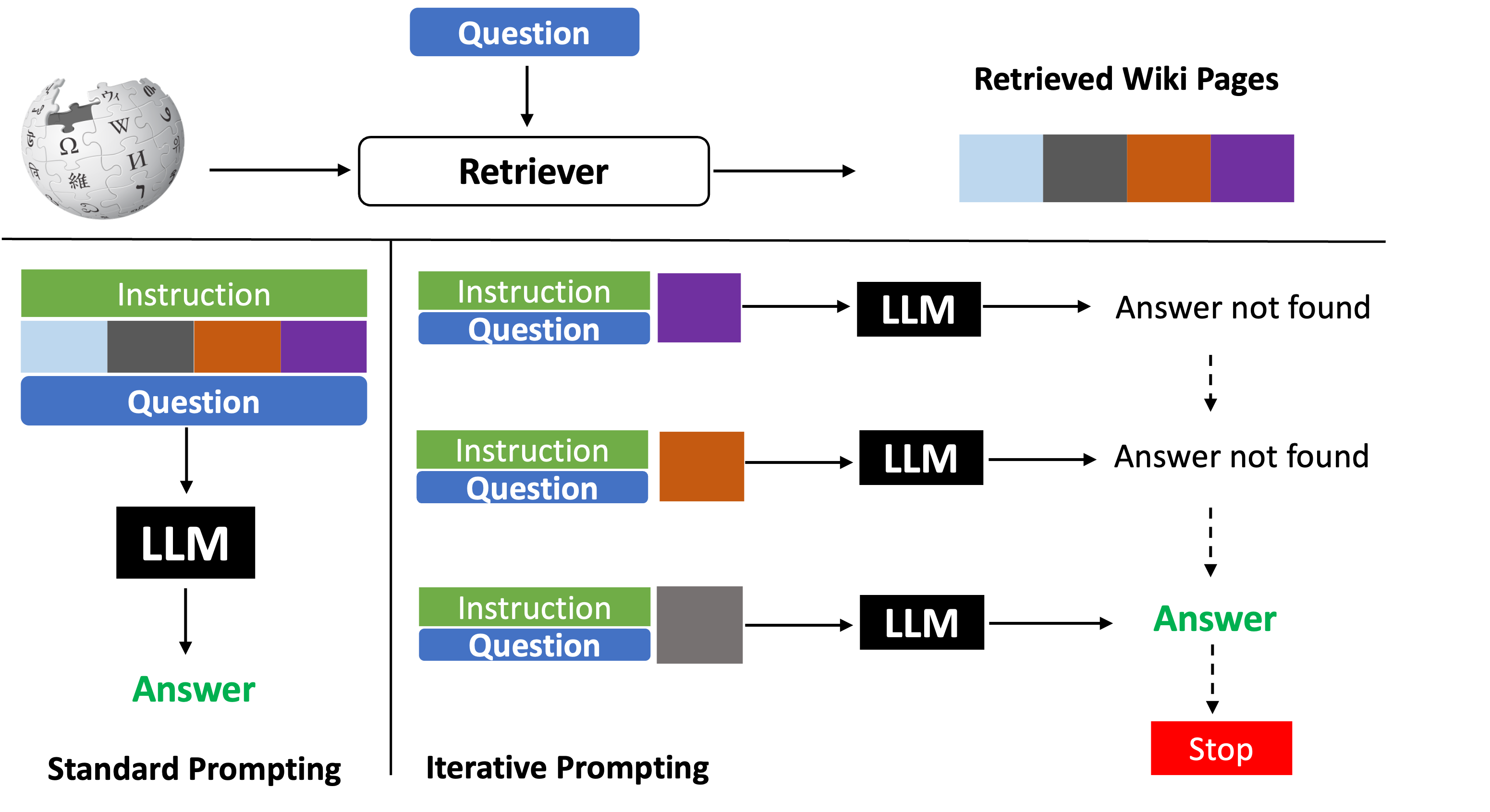}
    \caption{The pipeline of two  prompting strategies. Given a question, we first retrieve some relevant pages using an off-the shelf retriever~(the upper part). In the standard prompting (the lower left part), all retrieved pages along with the given question and a customized instruction, are fed to an LLM to generate the answer. In the adaptive prompting (illustrated in the lower right), we employ a sliding window approach. This window traverses the pages in descending order of retrieval score, sequentially feeding specific segments to the LLM to generate the answer.
    }
    \label{fig:pipeline}
\end{figure*}

Our experiments on three open-domain question answering datasets~\cite{kwiatkowski2019natural, yang2018hotpotqa, joshi2017triviaqa} confirm the efficacy of this approach. While matching or outperforming standard methods that utilize the full context at once, our strategy reduces token usage by over 50\% on average. This validates the dual benefit of improved computational efficiency and better noise resilience. Additionally, our ablation studies highlight the importance of the processing sequence. We find that LLMs tend to force incorrect answers from irrelevant passages rather than acknowledging a lack of information; thus, feeding high-probability context windows first is critical to prevent hallucinations.

\section{Method}
The overall iterative generation process is shown in Figure~\ref{fig:pipeline}. 
Given a question, we first retrieve the top-K Wikepedia pages.  
Unlike the conventional RAG system that fills the LLM context with all pages, our approach employs a sliding window which iterates through pages ranked by retrieval score, sequentially feeding them to the LLM for answer generation. 
If the current window of pages contains the correct answer, the LLM locates the position of the relevant page in the window, highlights the passage within the page that answers the question, and outputs the correct answer.  
Otherwise, the LLM provides a rationale for not providing an answer and proceeds to the subsequent window.

\subsection{Page Level Retrieval}
We use an in-house implementation of BM25 \cite{bm25} to retrieve top-K pages from Wikipedia
, where $K$ is a hyperparameter.
Compared to the dense retrieval method \cite{ni-etal-2022-large, wang2024textembeddingsweaklysupervisedcontrastive, lee2024geckoversatiletextembeddings, wang2024retrieve} which typically operates on fixed length passages, BM25 can encode arbitrarily long documents and allows us to build our retrieval index based on entire Wiki-pages rather than fixed length passages.
This avoids information loss caused by segmenting the document into passages~\cite{choi-etal-2021-decontextualization}, which was a non-negligible source of errors in our pilot studies. 

\subsection{Prompt Design}

\begin{figure*}[!t]
    \centering
    \includegraphics[width=0.96\linewidth,height=0.71\textheight]{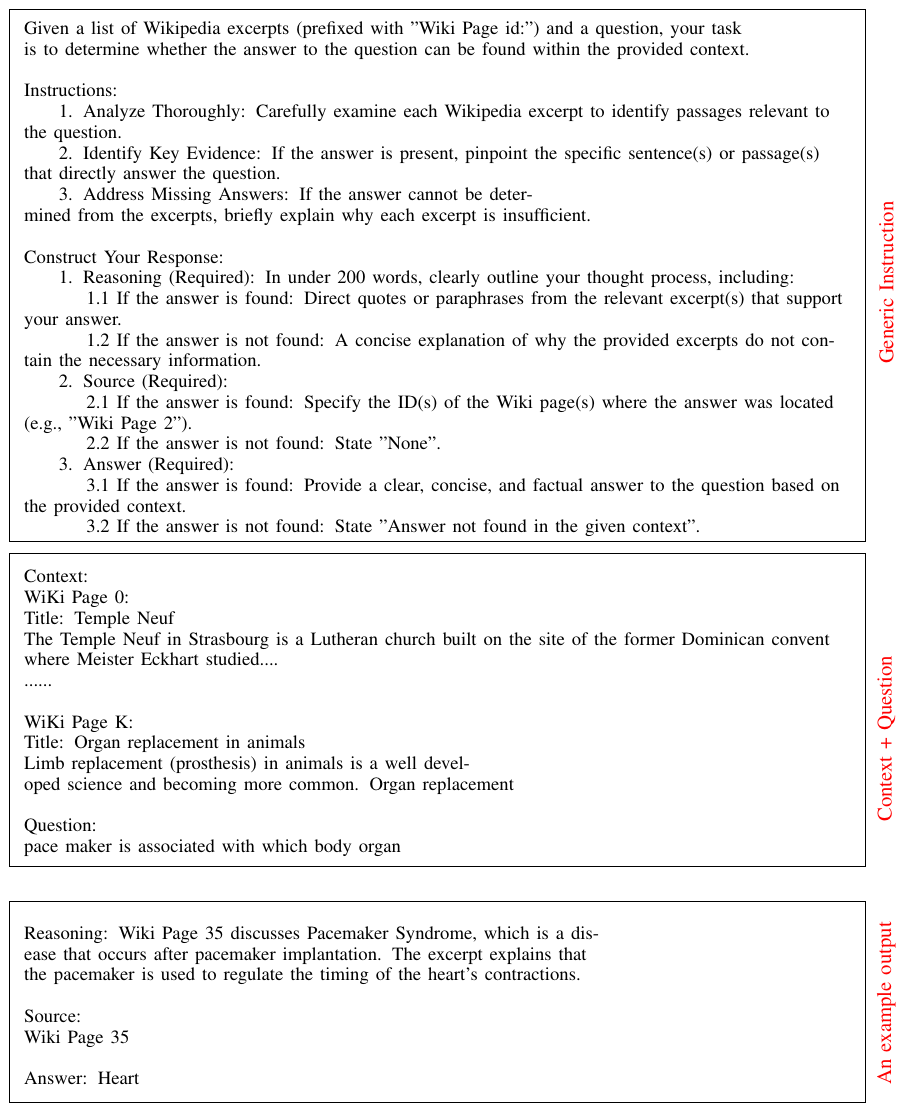}
    \caption{An example of zero-shot prompting}
    \label{fig:zero-shot}
\end{figure*}

We construct our prompt starting with a system-level instruction, a method proven to effectively align LLMs as assistants~\cite{lin2023unlockingspellbasellms}. This system prompt outlines the task description, details the reasoning process, and specifies the output format. The system instruction is followed by a window of $w$ Wikipedia pages (where $w < K$) and the target question. Specifically, each Wikipedia page is represented by: 1) page ID, 2) title, and 3) main content. An example is provided in Figure\ref{fig:zero-shot}. Within the system prompt, we follow prior work~\cite{azamfirei2023large,tonmoy2024comprehensive} by instructing the model to ground its answer solely in the provided context to mitigate hallucinations. Furthermore, by directing the LLM to locate the specific source page, we implicitly utilize the model as a reranker; requiring it to highlight relevant sentences further enhances interpretability.

\section{Experiments}
\begin{table*}[!t]
    \centering
    \begin{tabular}{lccccccccc}
      \toprule 
         &  R@1   &  R@5  &  R@10  &  R@20   &  R@50 &  R@100 &  R@200 &   R@500 & R@1000 \\ 
         \midrule 
        NQ & 21.84\ & 45.41\ & 55.72\ & 65.47\ & 75.79\ & 81.95 & 87.03 & 91.02 & 93.59 \\ 
         \midrule 
         HotPotQA &   24.91\ & 45.24\ & 52.63\ & 59.16\ & 67.12\ & 72.32\ & 76.99 & 82.68 &  86.12 \\
         \midrule 
         TriviaQA & 16.23\ & 38.71\ & 47.91\ & 56.78\ & 65.95\ & 72.15\ & 77.72\ & 83.35 & 87.29 \\

      \bottomrule 
    \end{tabular}
    \caption{BM25 Recall@K on NQ, HotPotQA and TriviaQA test set.}
    \label{tab:retrieve}
\end{table*}

We conduct experiments on three open-domain question-answering datasets from the KILT suite~\cite{petroni2020kilt}: Natural Questions (NQ)~\cite{kwiatkowski2019natural}, TriviaQA~\cite{joshi2017triviaqa}, and HotpotQA~\cite{yang2018hotpotqa}.
We utilize the 2019/08/01 Wikipedia dump curated by KILT as the external knowledge source.
Following \newcite{lu-etal-2024-hyrr} we implement BM25 based on WordPiece Tokenization, and fix K=0.9 and b=0.8.
We use the BERT~\textsubscript{base} \cite{devlin-etal-2019-bert} vocabulary with 30522 entries.
The test set recall metrics can be found in Table~\ref{tab:retrieve}.
All experiments reported in this paper are based on Gemini 1.5 Pro~\cite{reid2024gemini} with greedy decoding.
We choose Gemini 1.5 Pro for our experiments as it achieves state-of-the-art results on several benchmarks while offering the longest context length.

\begin{table*}[!t]
    \centering
    \setlength{\tabcolsep}{4.5pt} 
    \begin{tabular}{lcccccc}
    \toprule
         &  \multicolumn{2}{c}{NQ} & \multicolumn{2}{c}{TriviaQA} &  \multicolumn{2}{c}{HotpotQA}   \\ 
         \midrule  
         & EM & Avg. \# of Wiki pages   & EM & Avg. \# of Wiki pages & EM & Avg. \# of Wiki pages  \\ 
         \midrule 
        KILT & 48.8 & -- & 61.7 & -- & 27.7 & -- \\
        Top-50  & 55.9 & 50  & 82.9 & 50 & 50.2 & 50 \\
        Top-100  & 57.2 & 100  & \textbf{83.9} & 100 & 51.8 & 100 \\
        Top-200  & 56.9 & 200  & 83.1 & 200 & 51.1 & 200 \\
        Iterative  & \textbf{57.8} & 76.8 & \textbf{83.9} & 65.2 & \textbf{52.0} & 70.4 \\
    \bottomrule
    \end{tabular}
    \caption{Main results. Top-K denotes feeding all top-K Wiki pages to the LLM context.}
    \label{tab:main}
\end{table*}

\subsection{Zero-shot Results}

We compare our iterative prompting method against a standard baseline that ingests all top-$K$ pages at once. As shown in Table~\ref{tab:main}, our findings diverge from recent work~\cite{hsieh2024ruler, lee2024longcontextlanguagemodelssubsume} that posits longer contexts strictly harm performance due to noise. We observe an inverted-U pattern with fixed windows: performance improves when increasing window size from 50 to 100, but drops at 200. This indicates a tension between improved recall and increased noise accumulation.

By contrast, our iterative approach consistently achieves superior performance across all datasets. Moreover, it utilizes approximately 1.5$\times$ fewer tokens than the fixed-window method. This token efficiency translates directly to lower computational costs, making iterative prompting a more economical solution.

\subsection{Analysis}
We conduct ablation studies on Natural Questions to better understand the factors that affect the performance of the iterative prompting approach.

\begin{figure}[!htp]
    \begin{tikzpicture}[scale=0.38]
        \def\rectWidth{1.5}
        \def\rectHeight{1.3}
        \def\vertMargin{1.2}
        \def\rectMargin{0.2}
        \def\sHeight{0.4}
        \draw[] ( {5 * (\rectWidth + \rectMargin)}, 0) rectangle ++(\rectWidth, \rectHeight) node[midway]{1};
        \draw[] ( {6 * (\rectWidth + \rectMargin)}, 0) rectangle ++(\rectWidth, \rectHeight) node[midway]{2};
        \draw[draw=none] ( {7 * (\rectWidth + \rectMargin)}, 0) rectangle ++(\rectWidth, \rectHeight) node[midway]{\ldots};
        \draw[fill=green!20, draw=none] ( {8 * (\rectWidth + \rectMargin)}, 0) rectangle ++(\rectWidth, \rectHeight) node[midway]{m};
        \draw[draw=none] ( {9 * (\rectWidth + \rectMargin)}, 0) rectangle ++(\rectWidth, \rectHeight) node[midway]{\ldots};
        \draw[] ( {10 * (\rectWidth + \rectMargin)}, 0) rectangle ++(\rectWidth, \rectHeight) node[midway]{k};

        \draw[decorate,decoration={brace,amplitude=10pt,raise=4pt},yshift=1.6cm] 
            ({5 * (\rectWidth + \rectMargin)}, 0) -- ({10.8 * (\rectWidth + \rectMargin)}, 0) node[midway,yshift=0.6cm]{Positive Window};
            

         \draw[] ( {11.5 * (\rectWidth + \rectMargin)}, 1) rectangle ++(\rectWidth, \sHeight) node[midway, xshift=1.8cm]{an irrelevant page};

        \draw[fill=green!20] ( {11.5 * (\rectWidth + \rectMargin)}, 0.2) rectangle ++(\rectWidth, \sHeight) node[midway, xshift=2cm]{a relevant page };
       
    \end{tikzpicture}
  \caption{An example of positive window contain $k$ Wikipedia pages. The \textcolor{green}{green} rectangle represents a relevant Wikipedia page with respect to a given question.}
  \label{fig:positive-window}
\end{figure}
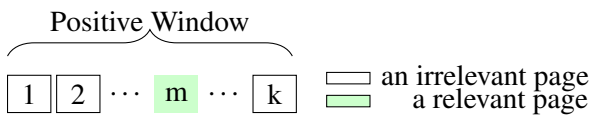
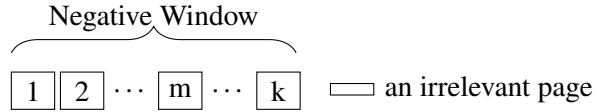
\begin{figure}[!htp]
    \begin{tikzpicture}[scale=0.38]
        \def\rectWidth{1.5}
        \def\rectHeight{1.3}
        \def\vertMargin{1.2}
        \def\rectMargin{0.2}
        \def\sHeight{0.4}
    
        \draw[] ( {5 * (\rectWidth + \rectMargin)}, 0) rectangle ++(\rectWidth, \rectHeight) node[midway]{1};
        \draw[] ( {6 * (\rectWidth + \rectMargin)}, 0) rectangle ++(\rectWidth, \rectHeight) node[midway]{2};
        \draw[draw=none] ( {7 * (\rectWidth + \rectMargin)}, 0) rectangle ++(\rectWidth, \rectHeight) node[midway]{\ldots};
        \draw[] ( {8 * (\rectWidth + \rectMargin)}, 0) rectangle ++(\rectWidth, \rectHeight) node[midway]{m};
        \draw[draw=none] ( {9 * (\rectWidth + \rectMargin)}, 0) rectangle ++(\rectWidth, \rectHeight) node[midway]{\ldots};
        \draw[] ( {10 * (\rectWidth + \rectMargin)}, 0) rectangle ++(\rectWidth, \rectHeight) node[midway]{k};

        \draw[decorate,decoration={brace,amplitude=10pt,raise=4pt},yshift=1.6cm] 
            ({5 * (\rectWidth + \rectMargin)}, 0) -- ({10.8 * (\rectWidth + \rectMargin)}, 0) node[midway,yshift=0.6cm]{Negative Window};
            

         \draw[] ( {11.5 * (\rectWidth + \rectMargin)}, 0.5) rectangle ++(\rectWidth, \sHeight) node[midway, xshift=1.8cm]{an irrelevant page};

       
    \end{tikzpicture}
    
    \caption{An example of negative window contain $k$ Wikipedia pages. It denotes that all $k$ Wikipedia pages are irrelevant to the question and the LLM should output ``answer not found'' when encountering such a window.}\label{fig:negative-window}
  
\end{figure}

\paragraph{Impact of Window Size} The window size governs the trade-off between recall and noise (Figure~\ref{fig:pipeline}). We label windows containing relevant pages as positive (see Figure~\ref{fig:positive-window})  and those without as negative (see Figure~\ref{fig:negative-window}). Larger windows are more likely to be positive but also introduce more distractors. Smaller windows reduce noise per step but require the model to process and correctly reject a longer sequence of negative windows, increasing the risk of cumulative error. As shown by the blue curve in Figure~\ref{fig:order} (left), a window size of 60 offers the best balance between the individual task difficulty and the total number of predictions.
\begin{figure} 
    \centering
\begin{tikzpicture}
    \begin{groupplot}[
        group style={
            group name=my plots,
            group size=2 by 1,
            horizontal sep=1.5cm,
        },
        width=4cm,
        height=4cm,
        xlabel style={font=\footnotesize},
        ylabel style={font=\footnotesize},
        xtick={1, 2, 3, 4},
        xticklabels={40, 60, 80, 200},
        legend style={font=\footnotesize, at={(0.5, -1.25)}, anchor=north, legend columns=-1},
        legend cell align={left},
        enlarge x limits=0.25,
    ]

    \nextgroupplot[
        ymin=25, ymax=65,
        ylabel=EM (\%),
        xlabel=Window size,
        ytick={25, 35, 45, 55, 65},
    ]
    \addplot+[black, mark=o] coordinates {(1, 56.2) (2, 57.5) (3, 57.1) (4, 56.9)};
    \addplot+[blue, mark=o] coordinates {(1, 34.2) (2, 32.8) (3, 41.2) (4, 55.4)};

    \nextgroupplot[
        ylabel={\# of negative windows},
        xlabel=Window size,
        ymin=0, ymax=10,
        ytick={0, 2, 4, 6, 8}
    ]
    \addplot+[black, mark=o] coordinates {(1, 0.7) (2, 0.4) (3, 0.2) (4, 0.1)};
    \addplot+[blue, mark=o] coordinates {(1, 8.3) (2, 3.8) (3, 1.9) (4, 0.9)};

    \end{groupplot}
    
    \node at ($(my plots c1r1.south east)!0.5!(my plots c2r1.south west) - (0, -2.7cm)$) {\begin{tabular}{@{}l@{}}
        \begin{tikzpicture}
            \draw[black, mark=o] (0,0) -- (0.5,0) node[right] {\footnotesize Forward};
            \draw[blue, mark=o] (3,0) -- (3.5,0) node[right] {\footnotesize Backward};
        \end{tikzpicture} 
    \end{tabular}};
    
\end{tikzpicture}
    \caption{Impact of the window sliding direction. The left figure illustrates the changes in EM across various window sizes when using forward and backward sliding directions. The right figure displays the corresponding average number of negative windows for each scenario.
    }
    \label{fig:order}
\end{figure}
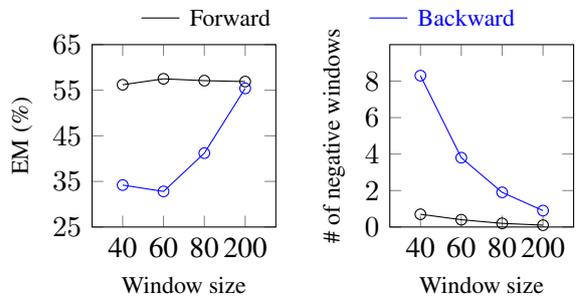

\begin{figure*}[!t]
    \begin{tikzpicture}[scale=0.98]
        \def\rectWidth{0.8}
        \def\rectHeight{0.6}
        \def\vertMargin{1.2}
        \def\rectMargin{0.2}
      \draw[] ( {0 * (\rectWidth + \rectMargin)},  {-3.2 * (\rectHeight  + \vertMargin)}) rectangle ++(\rectWidth, \rectHeight) node[midway] {1} ;
      
      \draw[] ( {1 * (\rectWidth + \rectMargin)},  {-3.2 * (\rectHeight  + \vertMargin)}) rectangle ++(\rectWidth, \rectHeight)node[midway,yshift=-0.7cm, xshift=0.6cm]{\small 
       1st Negative Window} node[midway] {2};
      
      \draw[] ( {2 * (\rectWidth + \rectMargin)},  {-3.2 * (\rectHeight  + \vertMargin)}) rectangle ++(\rectWidth, \rectHeight)  node[midway] {3} ;

     \draw[] ( {3 * (\rectWidth + \rectMargin)},  {-3.2 * (\rectHeight  + \vertMargin)}) rectangle ++(\rectWidth, \rectHeight) node[midway] {4};


      \draw[] ( {5 * (\rectWidth + \rectMargin)},  {-3.2 * (\rectHeight  + \vertMargin)}) rectangle ++(\rectWidth, \rectHeight) node[midway] {5};
      
      \draw[] ( {6* (\rectWidth + \rectMargin)},  {-3.2 * (\rectHeight  + \vertMargin)}) rectangle ++(\rectWidth, \rectHeight) node[midway,yshift=-0.7cm, xshift=0.6cm]{\small 2nd Negative Window} node[midway] {6};
      
      \draw[] ( {7* (\rectWidth + \rectMargin)},  {-3.2 * (\rectHeight  + \vertMargin)}) rectangle ++(\rectWidth, \rectHeight) node[midway] {7};

      \draw[] ( {8 * (\rectWidth + \rectMargin)},  {-3.2 * (\rectHeight  + \vertMargin)}) rectangle ++(\rectWidth, \rectHeight) node[midway] {8};


      \draw[] ( {10 * (\rectWidth + \rectMargin)},  {-3.2 * (\rectHeight  + \vertMargin)}) rectangle ++(\rectWidth, \rectHeight) node[midway] {9};
      
      \draw[] ( {11 * (\rectWidth + \rectMargin)},  {-3.2 * (\rectHeight  + \vertMargin)}) rectangle ++(\rectWidth, \rectHeight) node[midway,yshift=-0.7cm, xshift=0.6cm]{\small 3rd Negative Window} node[midway] {10};
      
      \draw[] ( {12* (\rectWidth + \rectMargin)},  {-3.2 * (\rectHeight  + \vertMargin)}) rectangle ++(\rectWidth, \rectHeight) node[midway] {11};

      \draw[] ( {13 * (\rectWidth + \rectMargin)},  {-3.2 * (\rectHeight  + \vertMargin)}) rectangle ++(\rectWidth, \rectHeight) node[midway] {12};


      \def\recH{2.2}

      \draw[] ( {0 * (\rectWidth + \rectMargin)},  {-3.2 * (\rectHeight  + \vertMargin) + \recH}) rectangle ++(\rectWidth, \rectHeight) node[midway] {1};

      \draw[] ( {1 * (\rectWidth + \rectMargin)},  {-3.2 * (\rectHeight  + \vertMargin) + \recH}) rectangle ++(\rectWidth, \rectHeight) node[midway] {2};
      
      \draw[] ( {2 * (\rectWidth + \rectMargin)},  {-3.2 * (\rectHeight  + \vertMargin) + \recH }) rectangle ++(\rectWidth, \rectHeight)  node[midway] {3} ;

     \draw[] ( {3 * (\rectWidth + \rectMargin)},  {-3.2 * (\rectHeight  + \vertMargin) + \recH}) rectangle ++(\rectWidth, \rectHeight) node[midway] {4};

       \draw[] ( {4 * (\rectWidth + \rectMargin)},  {-3.2 * (\rectHeight  + \vertMargin) + \recH}) rectangle ++(\rectWidth, \rectHeight) node[midway] {5};

      \draw[] ( {5 * (\rectWidth + \rectMargin)},  {-3.2 * (\rectHeight  + \vertMargin) + \recH }) rectangle ++(\rectWidth, \rectHeight) node[midway] {6};
      
      \draw[] ( {6* (\rectWidth + \rectMargin)},  {-3.2 * (\rectHeight  + \vertMargin) + \recH }) rectangle ++(\rectWidth, \rectHeight)  node[midway] {7};
      
      \draw[] ( {7* (\rectWidth + \rectMargin)},  {-3.2 * (\rectHeight  + \vertMargin) + \recH }) rectangle ++(\rectWidth, \rectHeight) node[midway] {8};

      \draw[] ( {8 * (\rectWidth + \rectMargin)},  {-3.2 * (\rectHeight  + \vertMargin) + \recH }) rectangle ++(\rectWidth, \rectHeight) node[midway] {9};

      \draw[] ( {9 * (\rectWidth + \rectMargin)},  {-3.2 * (\rectHeight  + \vertMargin) + \recH }) rectangle ++(\rectWidth, \rectHeight) node[midway] {10};

      \draw[] ( {10 * (\rectWidth + \rectMargin)},  {-3.2 * (\rectHeight  + \vertMargin) + \recH }) rectangle ++(\rectWidth, \rectHeight) node[midway] {11};
      
      \draw[] ( {11 * (\rectWidth + \rectMargin)},  {-3.2 * (\rectHeight  + \vertMargin) + \recH }) rectangle ++(\rectWidth, \rectHeight)  node[midway] {12};

      \draw[dotted, line width=0.8] ( {0 * (\rectWidth + \rectMargin)-0.1},  {-3.3 * (\rectHeight  + \vertMargin) + \recH}) rectangle ++(15 * \rectWidth, 1.6* \rectHeight);

        \draw[->, line width=2]  ( 1, -2) -- (1, -2.5) node[midway, right, xshift=0.4cm] {\small Gather all the pages that precede the first relevant page};

        \draw[->,line width=1]  ( 7, -4.2) -- (2, -5);
        \draw[->, line width=1]  ( 7, -4.2) -- (7, -5) node[midway, above, yshift=0.3cm] {\small Divide the collected pages into negative windows of the given size (=4 in this example)};
        \draw[->, line width=1]  ( 7, -4.2) -- (12, -5);

        \draw[fill=green!20, draw=none] (12.2, -3.1) rectangle ++(0.8 * \rectWidth, 0.6 * \rectHeight) node[midway,xshift=1.4cm]{\small a relevant page};
       \draw[] (12.2,-3.6) rectangle ++(0.8 * \rectWidth, 0.6 * \rectHeight) node[midway,xshift=1.46cm]{\small an irrelevant page };

       \def\recH{4.5}

    \draw[] ( {0 * (\rectWidth + \rectMargin)},  {-3.2 * (\rectHeight  + \vertMargin) + \recH}) rectangle ++(\rectWidth, \rectHeight) node[midway] {1};

      \draw[] ( {1 * (\rectWidth + \rectMargin)},  {-3.2 * (\rectHeight  + \vertMargin) + \recH}) rectangle ++(\rectWidth, \rectHeight) node[midway] {2};
      
      \draw[] ( {2 * (\rectWidth + \rectMargin)},  {-3.2 * (\rectHeight  + \vertMargin) + \recH }) rectangle ++(\rectWidth, \rectHeight)  node[midway] {3} ;

     \draw[] ( {3 * (\rectWidth + \rectMargin)},  {-3.2 * (\rectHeight  + \vertMargin) + \recH}) rectangle ++(\rectWidth, \rectHeight) node[midway] {4};

       \draw[] ( {4 * (\rectWidth + \rectMargin)},  {-3.2 * (\rectHeight  + \vertMargin) + \recH}) rectangle ++(\rectWidth, \rectHeight) node[midway] {5};

      \draw[] ( {5 * (\rectWidth + \rectMargin)},  {-3.2 * (\rectHeight  + \vertMargin) + \recH }) rectangle ++(\rectWidth, \rectHeight) node[midway] {6};
      
      \draw[] ( {6* (\rectWidth + \rectMargin)},  {-3.2 * (\rectHeight  + \vertMargin) + \recH }) rectangle ++(\rectWidth, \rectHeight)  node[midway] {7};
      
      \draw[] ( {7* (\rectWidth + \rectMargin)},  {-3.2 * (\rectHeight  + \vertMargin) + \recH }) rectangle ++(\rectWidth, \rectHeight) node[midway] {8};

      \draw[] ( {8 * (\rectWidth + \rectMargin)},  {-3.2 * (\rectHeight  + \vertMargin) + \recH }) rectangle ++(\rectWidth, \rectHeight) node[midway] {9};

      \draw[] ( {9 * (\rectWidth + \rectMargin)},  {-3.2 * (\rectHeight  + \vertMargin) + \recH }) rectangle ++(\rectWidth, \rectHeight) node[midway] {10};

      \draw[] ( {10 * (\rectWidth + \rectMargin)},  {-3.2 * (\rectHeight  + \vertMargin) + \recH }) rectangle ++(\rectWidth, \rectHeight) node[midway] {11};
      
      \draw[] ( {11 * (\rectWidth + \rectMargin)},  {-3.2 * (\rectHeight  + \vertMargin) + \recH }) rectangle ++(\rectWidth, \rectHeight)  node[midway] {12};
      
      \draw[draw=none, fill=green!20] ( {12* (\rectWidth + \rectMargin)},  {-3.2 * (\rectHeight  + \vertMargin) + \recH }) rectangle ++(\rectWidth, \rectHeight) node[midway] {13};

    \draw[draw=none] ( {13* (\rectWidth + \rectMargin)},  {-3.2 * (\rectHeight  + \vertMargin) + \recH }) rectangle ++(\rectWidth, \rectHeight) node[midway] {\dots};

    \draw[] ( {13.8* (\rectWidth + \rectMargin)},  {-3.2 * (\rectHeight  + \vertMargin) + \recH }) rectangle ++(\rectWidth, \rectHeight) node[midway] {200};

    \draw[line width=0.8] ( {0 * (\rectWidth + \rectMargin)-0.1},  {-3.3 * (\rectHeight  + \vertMargin) + \recH}) rectangle ++(18.7 * \rectWidth, 1.6* \rectHeight);
    \end{tikzpicture}

    \caption{Construction of negative windows for evaluation. We discard negative windows positioned after the first positive window (which contains the relevant page). This setting focuses the evaluation on the model's ability to refrain from answering irrelevant inputs before encountering the target information (positive window).}\label{fig:negative}

\end{figure*}
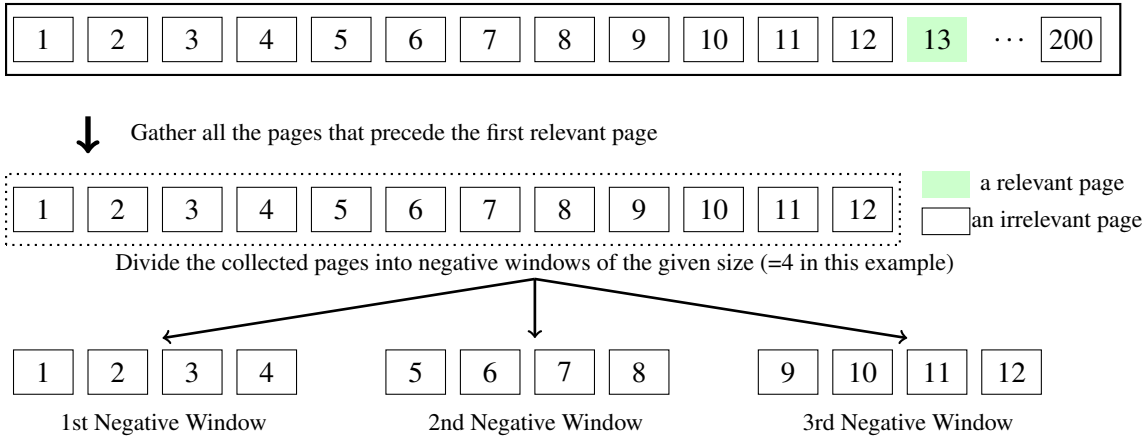

\paragraph{Impact of Sliding Window Order} To test robustness, we evaluate the system in a backward mode, sliding the window through pages sorted by ascending BM25 score. This simulates a scenario where relevant information is buried deep in the list of retrieved pages. Figure~\ref{fig:order} shows that while the standard (forward) approach encounters fewer than $0.5$ negative windows before seeing a positive window (at size 60), the backward approach encounters $3.8$. This forces the model to generate multiple "answer not found" predictions before reaching the target. As a result, the Exact Match (EM) score in the backward direction is considerably lower.

\paragraph{Difficulty of Negative Windows} The drop in performance in the backward setting suggests that the model struggles to handle negative windows. To quantify this, we evaluated the model solely on negative windows from NQ and TriviaQA. An example of constructing negative windows is shown in Figure~\ref{fig:negative}.  Since these windows contain no answer, the correct behavior is to output "answer not found".  However, we find that in a zero-shot setting, the model generates a false answer $54.3\%$ of the time, highlighting a high tendency toward hallucination when relevant context is absent.
%

Inspired by the efficacy of few-shot In-Context Learning (ICL) in enabling LLMs to adapt to new tasks, we investigate whether ICL can enhance the model's capacity for abstention i.e. the ability to decline to respond when information is insufficient. We evaluated performance across varying shot counts ($0, 2, 4, 6, 8$), maintaining a balanced distribution of positive and negative examples. Positive examples consisted of a sampled question, a positive window, and a target answer with a manually curated reasoning path. Negative examples consisted of a question, a negative window, and the target output ``answer not found''.

\begin{figure} 
    \centering
\begin{tikzpicture}
    \begin{groupplot}[
        group style={
            group name=my plots,
            group size=2 by 1,
            horizontal sep=1.6cm,
        },
        xlabel style={font=\footnotesize},
        ylabel style={font=\footnotesize},
        xtick={1, 2, 3, 4},
        xticklabels={40, 60, 80, 200},
        legend style={font=\footnotesize, at={(1, 1.25)}, anchor=north, legend columns=-1},
        legend cell align={left},
        enlarge x limits=0.25,
    ]
   \nextgroupplot[
        width=4cm,
        height=3.5cm,
        ymin=40, ymax=80,
        xmin=1, xmax=5,
        xtick={1, 2, 3, 4, 5},
        xticklabels={0, 2, 4, 6, 8},
        ylabel={Accuracy (\%)},
        xlabel={\# of shots},
        y label style={at={(axis cs:0,59.5)}},
        enlarge x limits=1.6,
        grid style={dotted, gray},
        grid= major,
        legend style={
            at={(1.27,1.4)},
            font=\small,
            legend columns=3,
            draw=none},
            axis x line=bottom,
            axis y line=left,
            /tikz/every even column/.append style={column sep=1.5cm}
        ]
        \addplot[thick, blue, mark=o, mark options={scale=1, fill=white}] coordinates {
            (1, 45.7)
            (2, 46.7)
            (3, 47.5)
            (4, 46.9)
            (5, 46.9)
        };
        \addlegendentry{NQ}

        \addplot[thick, black, mark=diamond, mark options={scale=1, fill=white}] coordinates {
            (1, 65.3)
            (2, 67.5)
            (3, 67.4)
            (4, 67.1)
            (5, 66.5)
        };
     \addlegendentry{TriviaQA}
   \nextgroupplot[
        width=4cm,
        height=3.5cm,
        ymin=40, ymax=100,
        xmin=1, xmax=5,
        xtick={1, 2, 3, 4, 5},
        xticklabels={0, 2, 4, 6, 8},
        ylabel={EM (\%)},
        y label style={at={(axis cs:0,63.5)}},
        xlabel={\# of shots},
        enlarge x limits=1.6,
        grid style={dotted, gray},
        grid= major,
        legend style={
        at={(1.07,1.)},
        font=\small,
        legend columns=3,
        draw=none},
        axis x line=bottom,
        axis y line=left,
        ]
        \addplot[thick, blue, mark=o, mark options={scale=1, fill=white}] coordinates {
            (1, 57.45)
            (2, 58.94)
            (3, 59.21)
            (4, 59.24)
            (5, 59.19)
        };

        \addplot[thick, black, mark=diamond, mark options={scale=1, fill=white}] coordinates {
            (1, 83.79)
            (2, 84.69)
            (3, 85.02)
            (4, 84.82)
            (5, 84.62)
        };
    
 \end{groupplot}
 \end{tikzpicture}
\caption{Accuracy of predicting the negative windows (left figure) and the corresponding exact match (right figure) using different numbers of shots.}
\label{tab:shots}
\end{figure}
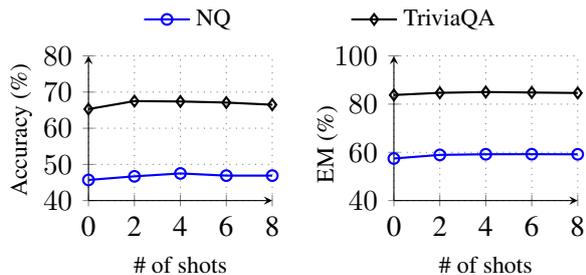

Figure~\ref{tab:shots} displays the impact of shot count on both accuracy of rejecting negative window and overall Exact Match (EM) scores. We observe that increasing the number of shots yields no significant improvement in either metric. This indicates that the model's struggle to identify negative windows is resistant to standard few-shot prompting. Unlike many other NLP tasks, providing more in-context examples is insufficient to teach the model to consistently decline to answer. These findings suggest that robust refusal capabilities may require intervention during training, such as fine-tuning with diverse instructional data or synthetic negative constraints.

\section{Related Work}

\paragraph{Retrieval-Augmented Generation}  
Large language models may produce factually incorrect responses to the given questions, as the knowledge embedded in the model parameters may be inaccurate, incomplete, and outdated~\cite{zhao2024retrieval}. To address the challenge,  Retrieval-Augmented Generation (RAG) has been introduced as a solution
~\cite{lewis2020retrieval,shuster2021retrieval,chen2024rulerag,chenc}. 
Unlike prior approaches~\cite{izacard2023atlas, levine2022standing, guu2020retrieval, lewis2020retrieval} 
that fine-tune models to adapt them to specific natural language processing
tasks,
this paper explores the increasingly prominent prompting-based methods~\cite{ram2023context, hsieh2024ruler, xuretrieval}, which have emerged as a mainstream paradigm for working with large language models (LLMs)~\cite{team2023gemini, bai2023qwen, meta2024introducing, gpt4}.

\paragraph{Evaluating Large Language Models}
Evaluating LLMs and identifying their limitations are essential steps towards driving their continued advancement~\cite{sun2023evaluating,guo2023evaluating,chang2024survey,zeng2023evaluating}. This is particularly important for long-context language models: although supported context lengths have improved dramatically - from early models such as GPT-3.5 with 4K tokens to recent models like Google Gemini 1.5 supporting up to 2 million tokens - numerous studies have shown that significant limitations remain. \citet{krishna2022rankgen} found that long-context neural generation in modestly-sized Transformer language models degrades due to the models' failure to properly condition on extended context.  More recently, \citet{liu2024lost} showed a U-shaped performance curve, indicating that the model's performance can degrade significantly when the position of relevant information is altered. \citet{hsiehruler,lee2024can} demonstrates that the effective context length can be substantially shorter than the claimed maximum context length. Different from prior work, we focus on RAG and design a sliding-window setting to investigate how LLMs respond to noisy retrieved passages, specifically whether they can admit ignorance at appropriate times instead of hallucinating. We argue that \textit{knowing when to admit ignorance} is a crucial capability that LLMs should possess.

\section{Conclusion and Discussion}
This study demonstrates that a divide-and-conquer approach to RAG can simultaneously improve retrieval accuracy and inference efficiency. By limiting the model's exposure to irrelevant information through a sliding window mechanism, we achieve superior performance compared to fixed-window baselines. However, our investigation highlights a persistent vulnerability in current LLMs: the struggle to process "negative windows."

Our results indicate that the success of iterative prompting depends heavily on the retrieval order. Because the model fails to abstain from answering when presented with irrelevant text - hallucinating false answers over 50\% of the time in zero-shot settings - placing relevant documents early in the sequence is critical to preventing errors. Crucially, we found that few-shot prompting (ICL) is ineffective at mitigating this behavior. This suggests that the "bias to answer" is deeply ingrained in current instruction-tuned models and cannot be easily overridden by prompt engineering alone. We conclude that while adaptive prompting is a powerful tool for managing context length, true robustness in RAG requires future research into training objectives that explicitly penalize unsupported generation and reward correct abstention.
\medskip 

\section{Limitations}
In this work, we conduct experiments with the Gemini Pro 1.5 model. We selected this architecture specifically because, at the time of this study, it supported the most extensive context window available and demonstrated state-of-the-art performance across diverse benchmarks. We acknowledge that the landscape of Large Language Models is rapidly evolving and that other models, such as the Gemini-3 family offers superior quality. While we did not perform an exhaustive cross-model evaluation, the primary objective of this research is to highlight a fundamental behavioral issue common to LLMs: their propensity to fabricate answers from insufficient context rather than correctly signaling a lack of information.


\bibliography{custom}

\onecolumn
\appendix








\end{document}